	\documentclass{article}

	\usepackage{geometry}
	\usepackage{amsmath}
	\usepackage{amssymb}

	\allowdisplaybreaks

	\usepackage[dvipsnames]{xcolor}
	
	\newcommand{\txt}[1]{\texttt{#1}}

	\usepackage[parfill]{parskip}
	
	\parskip = \baselineskip

	\usepackage{enumitem}

	\usepackage{url}

	\usepackage{stackengine}
	
	\usepackage{float}
	
	\floatstyle{plaintop}
	\newfloat{algorithm}{tbp}{alg}[section]
	\floatname{algorithm}{Algorithm}
	
	\usepackage{caption}
	\usepackage{tabularx}
	
	\usepackage{pseudo}
	
	\pseudodefinestyle{fullwidth}{
		begin-tabular =
		\tabularx{0.9\linewidth}{@{}
			r
			>{\pseudosetup}
			X
			>{\leavevmode\small\color{black!60}}
			p{0.3\linewidth}
			@{}},
		end-tabular = \endtabularx,
		setup-append = \pseudoeq
	}

	\usepackage{subcaption}

	\usepackage[hpos=0.5\paperwidth, vpos=0.05\paperheight]{draftwatermark}
	\SetWatermarkText{Approved for public release: distribution unlimited.}
	\SetWatermarkFontSize{0.25cm}
	\SetWatermarkColor{red}
	\SetWatermarkAngle{0}
	
	\usepackage{tikz}
	
		\usetikzlibrary{shapes.multipart}
		
		\usetikzlibrary{shapes}
		
		\usetikzlibrary{arrows.meta}
	
	\usepackage{booktabs}
	
	\usepackage{accents}
	\newcommand\tbar[1]{\accentset{\rule{.5em}{0.6pt}}{\accentset{\vspace{-0.05pt}}{#1}}}

	\newcommand{\bq}{\begin{equation}}
	\newcommand{\eq}{\end{equation}}

	\def\D{\mathbb{D}}
	\def\Dnn{\D_{ \not - }}
	\def\E{\mathbb{E}}
	\def\F{\mathcal{F}}
	\def\mh{\hat{\mu}}
	\def\O{\mathbb{O}}
	\def\Om{\Omega}
	\def\o{\omega}
	\def\R{\mathbb{R}}
	\def\S{\Sigma}
	\def\Sb{\tbar{\S}}
	
	\def\Ub{\tbar{U}}
	\def\Vb{\tbar{V}}

	\def\cx{\Theta}
	\def\cxi{\cx_i}
	\def\chxi{\hat{\cx}_i}

	\def\trans{^\mathsf{T}}

	\def\inv{^{-1}}

	\DeclareMathOperator{\rankop}{rank}
	
	\newcommand{\rank}[1]{\rankop({#1})}

	\DeclareMathOperator{\rowsop}{rows}
	
	\newcommand{\rows}[1]{\rowsop\left({#1}\right)}

	\DeclareMathOperator{\colsop}{cols}
	
	\newcommand{\cols}[1]{\colsop\left({#1}\right)}

	\DeclareMathOperator{\trace}{tr}
	
	\newcommand{\tr}[1]{\trace\left({#1}\right)}
	
	\DeclareMathOperator{\diag}{diag}
	
	\newcommand{\dg}[2]{\diag_{#1}\left({#2}\right)}

	\DeclareMathOperator{\sing}{sing}

	\newcommand{\subtext}[1]{_{ \text{#1} }}
	\newcommand{\suptext}[1]{^{ \text{#1} }}
	
	\def\th{\suptext{th}}
	\def\c{\suptext{\,cov}}
	\def\opt{\suptext{opt}}

\title{Covariance-Generalized Matching Component Analysis\\for Data Fusion and Transfer Learning}

\author{%
Nick Lorenzo\footnote{University of Dayton Research Institute, Dayton, OH, USA (\txt{Nicholas.Lorenzo@udri.udayton.edu})}\hspace{0.5in}%
Sean O'Rourke\footnote{Air Force Research Laboratory, Wright-Patterson Air Force Base, OH, USA (\txt{sean.orourke.3@us.af.mil})}\hspace{0.5in}%
Theresa Scarnati\footnote{Qualis Corporation, Huntsville, AL, USA (\txt{tscarnati@qualis-corp.com})}}

\date{August 5, 2022}

\begin{document}

\maketitle

\begin{abstract}\noindent
In order to encode additional statistical information in data fusion and transfer learning applications, we introduce a generalized covariance constraint for the matching component analysis (MCA) transfer learning technique.  We provide a closed-form solution to the resulting covariance-generalized optimization problem and an algorithm for its computation.  We call the resulting technique -- applicable to both data fusion and transfer learning -- covariance-generalized MCA (CGMCA).  We also demonstrate via numerical experiments that CGMCA is capable of meaningfully encoding into its maps more information than MCA.
\end{abstract}

\section{Introduction}

The matching component analysis (MCA) transfer learning technique was originally developed as a data augmentation strategy for building large, representative machine learning training sets within a data-limited environment \cite{MCA}.  Specifically, MCA maps a training domain and a testing domain into a low-dimensional, common domain using only a small number of matched train-test image pairs.  These maps minimize the expected distance between train-test image pairs within the common domain, subject to an identity matrix covariance constraint and an affine linear structure.  The training domain's optimal affine linear transformation -- encoded with information from the matched train-test image pairs -- is then applied to a large number of unmatched training images, resulting in a large number of common-domain image representations to be used as training inputs.

We are interested in extending the MCA application space to the fusion of data acquired from two different modalities.  By generalizing MCA's covariance constraint, we afford the meaningful encoding of additional statistical information into the MCA maps.  We note that this encoding may also be used in the original transfer learning context.

In this paper, we first mathematically develop the generalized covariance constraint.  We then provide an algorithm for its computation.  We also demonstrate via numerical experiments that the resulting technique -- which we call covariance-generalized MCA (CGMCA) -- is capable of encoding into its maps more information than MCA.

\section{Mathematical development of CGMCA}

Our development of the covariance-generalized optimization problem mirrors the development of the original MCA optimization problem \cite{MCA}, with some notation borrowed from \cite[p.~130]{Lorenzo_Dissertation}.

\subsection{Notation}

Let $a, b, c \in \mathbb{ N }$ with $c \le \min\{ a , b \}$, and let $v \in \R^c$.  Define
\bq
\dg{ a \times b }{ v } \in \R^{ a \times b }
\eq
to be the matrix whose first $c$ diagonal elements are the elements of $v$ (in order) and zero elsewhere.

Let
\begin{subequations}
\bq
\O^{ a \times b } \equiv \{ O \in \R^{ a \times b } \mid O\trans O = I_b \}
\eq
denote the set of $a \times b$ real matrices with orthonormal columns (this set is non-empty only if $b \le a$; when $b = a$, these matrices are orthogonal), let
\bq
\D_+^a \equiv \{ D \in \R^{ a \times a } \mid D = \dg{ a \times a }{ [ \delta_1 , \ldots , \delta_a ]\trans } ~ \exists ~ \delta_1 \ge \ldots \ge \delta_a > 0 \}
\eq
denote the set of diagonal $a \times a$ real invertible matrices with non-increasing elements, and let
\bq
\Dnn^{ a \times b } \equiv \{ D \in \R^{ a \times b } \mid D = \dg{ a \times b }{ [ \delta_1 , \ldots , \delta_{ \min\{ a , b \} } ]\trans } ~ \exists ~ \delta_1 \ge \ldots \ge \delta_{ \min\{ a , b \} } \ge 0 \}
\eq
denote the set of diagonal $a \times b$ real matrices with non-negative, non-increasing elements.
\end{subequations}

We reserve the undecorated but subscripted symbols $U_{ ( \, \cdot \, ) }$, $\S_{ ( \, \cdot \, ) }$, and $V_{ ( \, \cdot \, ) }$ and barred, subscripted symbols $\Ub_{ ( \, \cdot \, ) }$, $\Sb_{ ( \, \cdot \, ) }$, and $\Vb_{ ( \, \cdot \, ) }$ exclusively for use as factors in the SVD.  In particular, for a matrix $Z \in \R^{ \rows{ Z } \times \cols{ Z } }$, we denote an arbitrary but fixed thin SVD of $Z$ by
\begin{subequations}
\bq
\label{notn:thin_svd}
Z = U_Z \S_Z V_Z\trans, \qquad \text{where} \qquad U_Z \in \O^{ \rows{ Z } \times \rank{ Z } }, \quad \S_Z \in \D_+^{ \rank{ Z } }, \quad V_Z \in \O^{ \cols{ Z } \times \rank{ Z } },
\eq
and we denote a fixed full SVD of $Z$ by
\bq
\label{notn:full_svd}
Z = \Ub_Z \Sb_Z \Vb_Z\trans, \qquad \text{where} \qquad \Ub_Z \in \O^{ \rows{ Z } \times \rows{ Z } }, \quad \Sb_Z \in \Dnn^{ \rows{ Z } \times \cols{ Z } }, \quad \Vb_Z \in \O^{ \cols{ Z } \times \cols{ Z } },
\eq
\end{subequations}
with $\Ub_Z$ an arbitrary but fixed orthogonal completion of $U_Z$, with $\Vb_Z$ an arbitrary but fixed orthogonal completion of $V_Z$, and with $\Sb_Z$ having $\S_Z$ as a leading principal submatrix, with zeros elsewhere.

\subsection{Problem statement}%{Introduction}

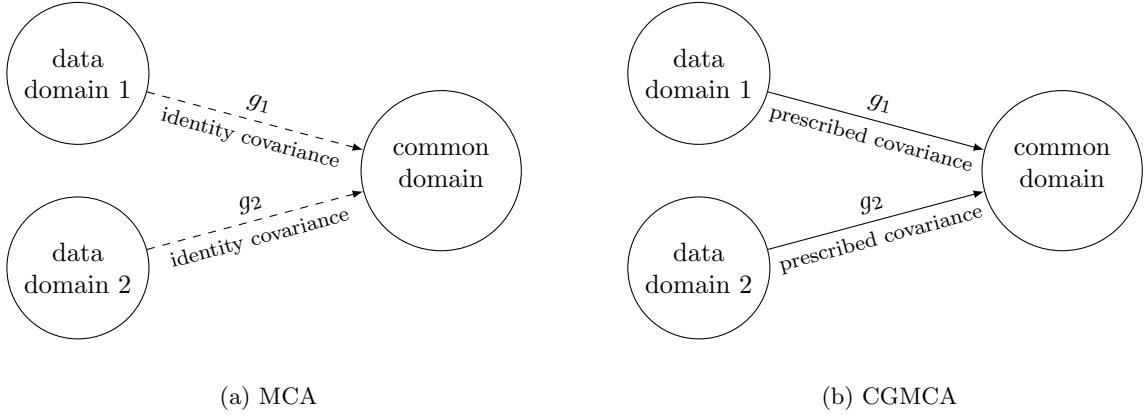
\begin{figure}[h]
\centering
\begin{subfigure}[t]{0.5\textwidth}\centering
	\begin{tikzpicture}[every text node part/.style={align=center}]
		\node[shape=circle,draw=black] (A) at (0,0) {\\{}common\\{}domain\\{}};
		\node[shape=circle,draw=black] (B) at (+165:5) {data\\ domain 1};
		\node[shape=circle,draw=black] (C) at (-165:5) {data\\ domain 2};
		\draw [dashed, -latex] (B) -- (A) node [midway, above, sloped] () {$g_1$} node [midway, below, sloped] () {\footnotesize{identity covariance}};
		\draw [dashed, -latex] (C) -- (A) node [midway, above, sloped] () {$g_2$} node [midway, below, sloped] () {\footnotesize{identity covariance}};
	\end{tikzpicture}
	\vspace{0.15in}
	\caption{MCA}
\label{fig:cov_mca}
\end{subfigure}%
\begin{subfigure}[t]{0.5\textwidth}\centering
	\begin{tikzpicture}[every text node part/.style={align=center}]
		\node[shape=circle,draw=black] (A) at (0,0) {\\{}common\\{}domain\\{}};
		\node[shape=circle,draw=black] (B) at (+165:5) {data\\ domain 1};
		\node[shape=circle,draw=black] (C) at (-165:5) {data\\ domain 2};
		\draw [-latex] (B) -- (A) node [midway, above, sloped] () {$g_1$} node [midway, below, sloped] () {\footnotesize{prescribed covariance}};
		\draw [-latex] (C) -- (A) node [midway, above, sloped] () {$g_2$} node [midway, below, sloped] () {\footnotesize{prescribed covariance}};
	\end{tikzpicture}
	\vspace{0.15in}
	\caption{CGMCA}
\label{fig:cov_cg}
\end{subfigure}
\caption{Schematic representation of the covariance constraints of MCA and CGMCA.}
\label{fig:cov_schematic}
\end{figure}

We consider random variables $X_1$ and $X_2$ with domain probability space $( \Om , \mathcal{ M } , P )$ and a set of outcomes $\{ \o_j \}_{ j \in [ n ] } \subseteq \Om$ for some $n \in \mathbb{ N }$ with $1 < n$, and we are provided, for $i \in \{ 1 , 2 \}$, with the data set $\{ X_i( \o_j ) \}_{ j \in [ n ] }$ of independent random variates belonging to the $i\th$ data domain $\Om_{ X_i } = \R^{ d_i }$ for some $d_i \in \mathbb{ N }$ with $1 < d_i$.  For each $j \in [ n ]$, we refer to the paired realizations $( X_1( \o_j ) , X_2( \o_j ) )$ as a matched data point.  Provided a covariance matrix $C_i C_i\trans \in \R^{ k \times k }$ for each $i \in \{ 1 , 2 \}$ for some fixed $k \in \mathbb{ N }$, we wish to use the matched data to approximately solve the constrained optimization problem
%\comment{See (C.10) of your dissertation for why this notation is okay (appearing to overload $g_i$).}
\begin{subequations}
\label{original_opt_problem}
\begin{align}
\label{original_obj_fcn}
\text{minimize} \quad & \E\| g_1( X_1 ) - g_2( X_2 ) \|_2^2,\\
\label{original_constraint_mem}
\text{subject to} \quad & g_i : \R^{ d_i } \to \R^k,\\
\label{original_constraint_ctr}
 & \E \, g_i( X_i ) = 0,\\
\label{original_constraint_cov}
 & \E\left[ g_i( X_i ) g_i( X_i )\trans \right] = C_i C_i\trans,\\
 \nonumber%
 & i \in \{ 1 , 2 \}.
\end{align}
\end{subequations}
The optimization problem \eqref{original_opt_problem} is our covariance-generalized version of the optimization problem (2.1) of \cite{MCA} (in which $C_i C_i\trans \equiv I_k$); we seek to minimize the expected distance between the images under $g_i$ of the random variables $X_i$ within the common domain of dimension $k$, while centering the resulting mappings $g_i$ via \eqref{original_constraint_ctr} and preserving a prescribed covariance structure via \eqref{original_constraint_cov}.  Figure \ref{fig:cov_schematic} illustrates schematically the difference between the covariance constraints of MCA and CGMCA.

\subsection{Approximation via available data}

Restricting $g_i$ to be an affine linear transformation, the requirement \eqref{original_constraint_mem} is equivalent to the requirement that $g_i( x ) = A_i x + b_i$ for some $A_i \in \R^{ k \times d_i }$ and $b_i \in \R^k$.  Denoting\footnote{As compared to (2.3) of \cite{MCA}, we have swapped the notations $\mu_i \leftrightarrow \mh_i$.} by $\mh_i$ and $\chxi$ the mean and covariance, respectively, of $X_i$, the centering constraint \eqref{original_constraint_ctr} then implies the condition $b_i = -A_i \mh_i$, so that
\bq
\label{g_i_form}
g_i( x ) = A_i ( x - \mh_i ).
\eq
The covariance constraint \eqref{original_constraint_cov} then becomes the condition $A_i \chxi A_i\trans = C_i C_i\trans$, and the optimization problem \eqref{original_opt_problem} becomes the optimization problem
\begin{subequations}
\label{opt_1}
\begin{align}
\text{minimize} \quad & \E\| A_1 ( X_1 - \mh_1 ) - A_2 ( X_2 - \mh_2 ) \|_2^2,\\
\text{subject to} \quad & A_i \in \R^{ k \times d_i },\\
	& A_i \chxi A_i\trans = C_i C_i\trans,\\
\nonumber%
	& i \in \{ 1 , 2 \}.
\end{align}
\end{subequations}
Using the $n$ realizations of $X_1$ and $X_2$ to which we have access, we approximate the true means $\mh_i$ and true covariances $\chxi$ via the sample means $\mu_i$ and sample covariances $\cxi$:
\begin{subequations}
\begin{align}
\label{mu_i_defn}
\mh_i \approx \mu_i & \equiv \frac{ 1 }{ n } \sum_{ j \in [ n ] } X_i( \o_j ) \in \R^{ d_i },\\
\label{Sigma_i_defn}
\chxi \approx \cxi & \equiv S_i S_i\trans \in \R^{ d_i \times d_i },
\end{align}
\end{subequations}
where we define $S_i \in \R^{ d_i \times n }$ to be the matrix whose $j\th$ column is the vector
\bq
\label{data_mat_col}
( S_i )_j \equiv \frac{ 1 }{ \sqrt{ n - 1 } } ( X_i( \o_j ) - \mu_i ) \in \R^{ d_i }.
\eq
Then the optimization problem \eqref{opt_1} is approximated by the optimization problem
\begin{subequations}
\begin{align}
\text{minimize} \quad & \frac{ n - 1 }{ n } \sum_{ j \in [ n ] } \left\| A_1 ( S_1 )_j - A_2 ( S_2 )_j \right\|_2^2,\\
\text{subject to} \quad & A_i \in \R^{ k \times d_i },\\
	& A_i \cxi A_i\trans = C_i C_i\trans,\\
\nonumber%
	& i \in \{ 1 , 2 \},
\end{align}
\end{subequations}
which is equivalent to the optimization problem
\begin{subequations}
\label{opt_2}
\begin{align}
\label{opt_2_obj_fcn}
\text{minimize} \quad & \frac{ n - 1 }{ n } \| A_1 S_1 - A_2 S_2 \|_F^2,\\
\text{subject to} \quad & A_i \in \R^{ k \times d_i },\\
\label{opt_2_constraint_cov}
	& A_i \cxi A_i\trans = C_i C_i\trans,\\
\nonumber%
	& i \in \{ 1 , 2 \}.
\end{align}
\end{subequations}

\subsection{Change of variables}

We make the change of variables
\bq
\label{A_i_defn}
A_i = U_{ C_i } \S_{ C_i } D_i \S_{ S_i }\inv U_{ S_i }\trans,
\eq
where the semi-orthogonal matrix
\bq
\label{D_i}
D_i\trans \in \O^{ r_i \times r_i\c }
\eq
is unknown.  With $A_i$ defined as in \eqref{A_i_defn}, the covariance constraint \eqref{opt_2_constraint_cov} is satisfied for each $D_i$ satisfying \eqref{D_i}.  Here,
\begin{subequations}
\bq
r_i \equiv \rank{ S_i } = \rank{ \cxi }
\eq
and
\bq
\label{r_i_cov_defn}
r_i\c \equiv \rank{ C_i } = \rank{ C_i C_i\trans } \le k.
\eq
\end{subequations}
We find that the covariance constraint \eqref{opt_2_constraint_cov} implies infeasibility unless
\bq
\label{feasibility_requirement}
r_i \ge r_i\c;
\eq
since $\cxi$ is a sample covariance matrix, $r_i$ always satisfies $r_i \le \min\{ d_i , n - 1 \}$.

\subsection{Transformation to trace maximization problem}

Using \eqref{A_i_defn}, we find that
\bq
A_i S_i = U_{ C_i } \S_{ C_i } D_i V_{ S_i }\trans,
\eq
so that the norm in the objective function \eqref{opt_2_obj_fcn} becomes
\begin{subequations}
\begin{align}
\| A_1 S_1 - A_2 S_2 \|_F^2
	& = \tr{ ( A_1 S_1 ) ( A_1 S_1 )\trans } + \tr{ ( A_2 S_2 ) ( A_2 S_2 )\trans } - 2 \tr{ ( A_1 S_1 ) ( A_2 S_2 )\trans }\\
	& = \tr{ C_1 C_1\trans + C_2 C_2\trans } - 2 \tr{ U_{ C_1 } \S_{ C_1 } D_1 V_{ S_1 }\trans V_{ S_2 } D_2\trans \S_{ C_2 } U_{ C_2 }\trans }.
\end{align}
\end{subequations}
The optimization problem \eqref{opt_2} can then be solved by finding
\begin{subequations}
\label{D_opt_problem}
\begin{align}
( D_1\opt , D_2\opt )
	& \in \arg \max_{ ( D_1 , D_2 ) \in \F } \tr{ U_{ C_1 } \S_{ C_1 } D_1 V_{ S_1 }\trans V_{ S_2 } D_2\trans \S_{ C_2 } U_{ C_2 }\trans }\\
	\label{simple_trace}
	& = \arg \max_{ ( D_1 , D_2 ) \in \F } \tr{ D_2\trans A\trans D_1 B },
\end{align}
where
\bq
\label{opt_trace_soln_A}
A \equiv \S_{ C_1 } U_{ C_1 }\trans U_{ C_2 } \S_{ C_2 } \quad\text{and}\quad B \equiv V_{ S_1 }\trans V_{ S_2 },
\eq
and where
\bq
\F \equiv \{ ( D_1 , D_2 ) \mid D_i \text{ satisfies \eqref{D_i}} , \, i \in \{ 1 , 2 \} \}
\eq
\end{subequations}
is the set of doublets of feasible matrices.

\subsection{A closed-form solution to the trace maximization problem}

The trace of \eqref{simple_trace} satisfies the bound \cite{tenBerge}
\bq
\tr{ D_2\trans A\trans D_1 B } \le \sum_{ j \in [ r_- ] } \sing_j( A ) \sing_j( B ),
\eq
where $r_- \equiv \min\{ \rank{ A } , \rank{ B } \}$.  This bound is achieved for
\bq
\label{D_i_opt}
( D_1\opt , D_2\opt ) = \left( \Ub_A \left( \left( \Ub_B \right)_{ [ r_1\c ] } \right)\trans , \Vb_A \left( \left( \Vb_B \right)_{ [ r_2\c ] } \right) \trans \right),
\eq
with $( \, \cdot \, )_{ [ \ell ] }$ denoting the first $\ell$ columns of a matrix; this solution is a semi-orthogonal extension of the solution provided in \cite[p.~463]{MatrixAnalysis}.

\subsection{Reduction of our optimal solution in the case of an identity matrix covariance constraint}

For the special case $C_i C_i\trans = I_k$ for $i \in \{ 1 , 2 \}$ investigated in \cite{MCA}, we find that $A = I_k$ in \eqref{opt_trace_soln_A}, so that our solution \eqref{D_i_opt} reduces to
\bq
( D_1\opt , D_2\opt ) = \left( \left( \Ub_B^{ [ k ] } \right)\trans , \left( \Vb_B^{ [ k ] } \right)\trans \right),
\eq
which is the solution provided in \cite{MCA}.

\section{Main results}

Here we state our main results.

\subsection{CGMCA maps}

Substituting the solution \eqref{D_i_opt} into the definition \eqref{A_i_defn} and using the approximation \eqref{mu_i_defn} in the result \eqref{g_i_form}, we find affine linear transformations approximately solving the original optimization problem \eqref{original_opt_problem} to be given by
\begin{subequations}
\label{opt_soln}
\bq
g_i\suptext{CGMCA}( x ) \equiv A_i\suptext{CGMCA} x + b_i\suptext{CGMCA},
\eq
where
\bq
A_i\suptext{CGMCA} \equiv U_{ C_i } \S_{ C_i } D_i\opt \S_{ S_i }\inv U_{ S_i }\trans, \quad
b_i\suptext{CGMCA} \equiv -A_i\suptext{CGMCA} \mu_i, \quad
i \in \{ 1 , 2 \},
\eq
\end{subequations}
and where $A$ and $B$ of the solutions $\{ D_i\opt \}$ of \eqref{D_i_opt} are defined as in \eqref{opt_trace_soln_A}.  The affine linear transformations $\{ g_i\suptext{CGMCA} \}$ are the CGMCA maps.

\subsection{CGMCA algorithm}

In Algorithm \ref{CovGenMCA}, we provide a method to compute the CGMCA maps \eqref{opt_soln} from the data and prescribed covariance matrices.

\begin{algorithm}[H]
\centering
\begin{pseudo}[fullwidth, line-height=1.25]*
\toprule
	\multicolumn{2}{l}{\pseudopr{Covariance-Generalized MCA}}
	&
	\\[bol=\midrule]
	\multicolumn{2}{l}{\kw{Inputs:} data $\{ X_i( \o_j ) \}_{ j \in [ n ] }$ and prescribed covariance matrices $C_i C_i\trans \in \R^{ k \times k }$ for $i \in \{ 1 , 2 \}$}
	\\
	\kw{for} $i \in \{ 1 , 2 \}$
	&
	\\+
		Compute sample mean $\mu_i$
		& See \eqref{mu_i_defn}
		\\
		Compute scaled and centered data matrix $S_i$
		& See \eqref{data_mat_col}
		\\
		Compute thin SVD $S_i = U_{ S_i } \S_{ S_i } V_{ S_i }\trans$
		& See \eqref{notn:thin_svd}
		\\
		Compute thin SVD $C_i C_i\trans = U_{ C_i } \S_{ C_i }^2 U_{ C_i }\trans$
		& See \eqref{notn:thin_svd}
		\\
		\kw{if} $r_i < r_i\c$
		&
		\\+
			\kw{return error}
			& Problem infeasible; see \eqref{feasibility_requirement}
			\\-
		\kw{end}
		&
		\\
		Compute matrix $\S_{ C_i } = ( \S_{ C_i }^2 )^{ 1 / 2 }$
		&
		\\-
	\kw{end}
	&
	\\
	Compute matrices $A = \S_{ C_1 } U_{ C_1 }\trans U_{ C_2 } \S_{ C_2 }$ and $B = V_{ S_1 }\trans V_{ S_2 }$
	& See \eqref{opt_trace_soln_A}
	\\
	Compute full SVDs $A = \Ub_A \Sb_A \Vb_A\trans$ and $B = \Ub_B \Sb_B \Vb_B\trans$
	& See \eqref{notn:full_svd}
	\\
	\kw{for} $i \in \{ 1 , 2 \}$
	&
	\\+
		Compute solution $D_i\opt$
		& See \eqref{D_i_opt}
		\\
		Compute linear map $A_i\suptext{CGMCA} = U_{ C_i } \S_{ C_i } D_i\opt \S_{ S_i }\inv U_{ S_i }\trans$
		& See \eqref{opt_soln}
		\\
		Compute translation $b_i\suptext{CGMCA} = -A_i\suptext{CGMCA} \mu_i$
		& See \eqref{opt_soln}
		\\-
	\kw{end}
	&
	\\*
\bottomrule
\end{pseudo}
\caption{Covariance-Generalized MCA.}
\label{CovGenMCA}
\end{algorithm}

\section{Numerical experiments}

In these numerical experiments, we demonstrate the claim that CGMCA is capable of meaningfully encoding into its maps more information than MCA.

\subsection{Data domains}

\begin{figure}[h!]
\centering
\includegraphics[scale=0.2]{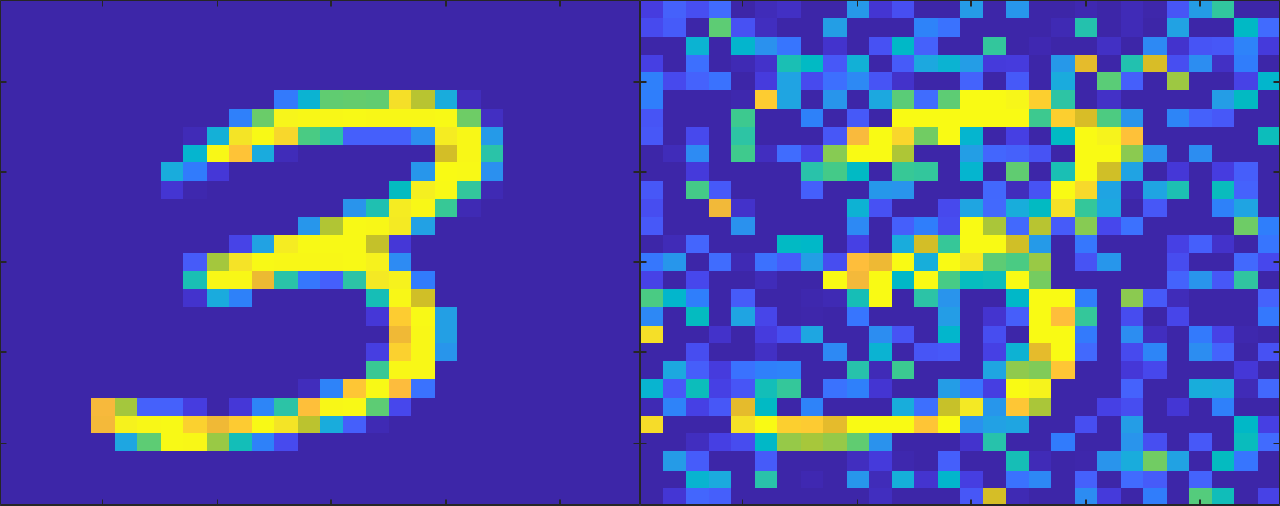}
\caption{An example of an unmodified (left) and corrupted (right) ``3''.}
\label{fig:data_domain_montage}
\end{figure}

%\comment{Careful here: how are we using $n$?  Is it consistent with how it's used previously in the paper?  Previously, it was for the matched data points.  Later, we partition $[ n ]$ to form train and test indices, which is not consistent with this idea.  So maybe use $N$ instead of $n$.  We then have $n_d \equiv | \text{inds}\subtext{train}^{ ( d ) } |$ as the number of matched data points in the map training for digit $d$.  Then $N = 60,000$ because our expt used all the MNIST digits.}
%
Using the MNIST \cite{MNIST} data set, we define two data domains.  The first is the set $\{ X\subtext{unmodified}( \o_j ) \}_{ j \in [ N ] }$ of unmodified MNIST digits, while the second is the set $\{ X\subtext{corrupted}( \o_j ) \}_{ j \in [ N ] }$ of MNIST digits corrupted with additive Gaussian noise with mean $0$ and standard deviation $0.1$.  Here, for each $j$, $X\subtext{unmodified}( \o_j ) \in \R^{ 784 \times 1 }$ and $X\subtext{corrupted}( \o_j ) \in \R^{ 784 \times 1 }$ are vectorized versions of the $( 28 \times 28 )$-pixel MNIST images and their corrupted counterparts.  An image from each of these domains is found in Figure \ref{fig:data_domain_montage}.  We also note that $N = 60,000$ is the total number of MNIST images.

\subsection{MCA and CGMCA map training}\label{sec:training}

For each digit $d \in \{ 0, \ldots, 9 \}$, define the set
\bq
\label{Xc_unmodified_d}
\mathcal{ X }\subtext{unmodified}^{ ( d ) } \equiv \{ X\subtext{unmodified}( \o_j ) \mid \text{label}( X\subtext{unmodified}( \o_j ) ) = d \}
\eq
of unmodified MNIST images of the digit $d$, and define the set $\mathcal{ X }\subtext{corrupted}^{ ( d ) }$ similarly.  In each experiment, one particular digit $d$ is chosen, and the set
\bq
\text{inds}_d \equiv \{ j \in [ N ] \mid X\subtext{unmodified}( \o_j ) \in \mathcal{ X }\subtext{unmodified}^{ ( d ) } \} = \{ j \in [ N ] \mid X\subtext{corrupted}( \o_j ) \in \mathcal{ X }\subtext{corrupted}^{ ( d ) } \}
\eq
of indices of images of the digit $d$ is partitioned into a set $\text{inds}\subtext{train}^{ ( d ) }$ of training indices and a set $\text{inds}\subtext{test}^{ ( d ) }$ of test indices, with the training indices comprising 80\% of the indices in $\text{inds}_d$ and the test set comprising the remaining 20\%.  Define the set
\bq
\mathcal{ X }\subtext{unmodified, train}^{ ( d ) } \equiv \{ X\subtext{unmodified}( \o_j ) \}_{ j \in \text{inds}\subtext{train}^{ ( d ) } },
\eq
with $\mathcal{ X }\subtext{corrupted, train}^{ ( d ) }$ defined similarly.  The data matrices whose columns are the elements of $\mathcal{ X }\subtext{unmodified, train}^{ ( d ) }$ and whose elements are the columns of $\mathcal{ X }\subtext{corrupted, train}^{ ( d ) }$ are the data matrices that serve as inputs to CGMCA in Algorithm \ref{CovGenMCA}.  The number of matched data points for the application of MCA and CGMCA for the digit $d$, then, is $n_d \equiv | \text{inds}\subtext{train}^{ ( d ) } |$.

The covariance matrices for CGMCA (that is, the inputs for Algorithm \ref{CovGenMCA}) were chosen to be the best approximation of rank $t$ of the covariance of the unmodified training data matrix whose columns are the elements of the set $\mathcal{ X }\subtext{unmodified, train}^{ ( d ) }$.  Here, $t$ is a parameter to be chosen.  For CGMCA, then, $k\subtext{CGMCA} = 784$, while $( r_i\c )\subtext{CGMCA} = t$.  For MCA, however, $t = ( r_i\c )\subtext{MCA} \equiv k\subtext{MCA}$.  The covariance matrices for the two techniques, then, have different sizes (values of $k$) but identical ranks (values of $r_i\c$).  In Table \ref{tab:ssim}, we show the values of $t$ chosen for each experiment.

This choice of covariance matrix for CGMCA reflects the goal of the testing procedure of Section \ref{sec:test}.  There, we wish to take as input a corrupted image $X\subtext{corrupted}( \o_j )$ for some $j$ and find its unmodified counterpart $X\subtext{unmodified}( \o_j )$; assigning -- via CGMCA -- the covariance in the common domain to be an approximation of the covariance of the unmodified images, then, allows an inversion process (chosen to be least squares in Section \ref{sec:test}) to simply preserve the common-domain covariance.  This is in contrast to MCA, where the identity covariance enforced in the common domain must be transformed by the inversion process into a more appropriate covariance (that of the unmodified data points).  See Section \ref{sec:results} for additional comments.

The maps $g\subtext{unmodified}\suptext{(CG)MCA}$ and $g\subtext{corrupted}\suptext{(CG)MCA}$ of \eqref{opt_soln} are then trained on the data $\mathcal{ X }\subtext{unmodified, train}^{ ( d ) }$ and $\mathcal{ X }\subtext{corrupted, train}^{ ( d ) }$, respectively; these data serve as the inputs for Algorithm \ref{CovGenMCA}, while the resulting maps are the outputs computed by Algorithm \ref{CovGenMCA}.

\subsection{Testing procedure}\label{sec:test}

After the MCA and CGMCA maps are trained as in Section \ref{sec:training}, we compare the performance of both techniques using the following testing procedure.

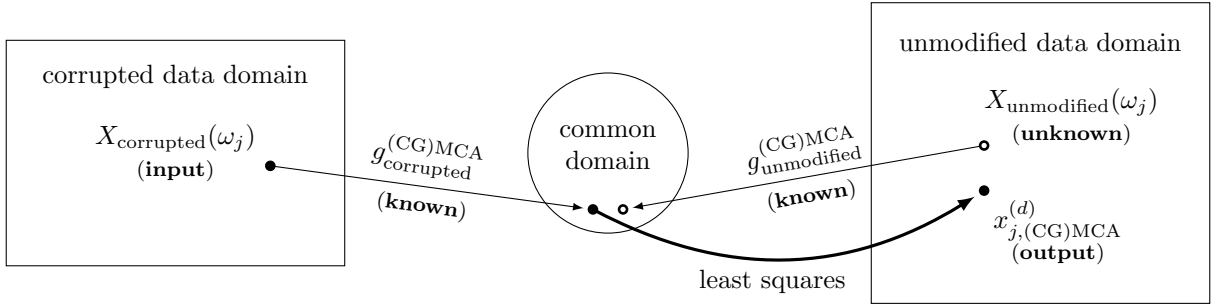
\begin{figure}[h]
\centering
	\begin{tikzpicture}[every text node part/.style={align=center}]
		\node[shape=circle,draw=black] (A) at (0,0) {\\{}common\\{}domain\\{}};
		\fill (-0.2,-0.75) circle (2pt) node (PL) {};
		\fill (+0.2,-0.75) circle (2pt) node (PR) {};
%		\fill (+5,+0.5) circle (2pt) node (U) [anchor=south west] {$X\subtext{unmodified}( \o_j )$\\\footnotesize{(\textbf{unknown})}};
		\fill[fill = white] (+0.2,-0.75) circle (1pt);
		\node[inner sep = 0pt] (U) [anchor=south west] at (+5,+0.1) {$X\subtext{unmodified}( \o_j )$\\\footnotesize{(\textbf{unknown})}};
		\fill (+5,-0.5) circle (2pt) node (O) [anchor=north west] {$x_{ j , \text{(CG)MCA} }^{ ( d ) }$\\\footnotesize{(\textbf{output})}};
		\draw[draw=black] (-8,-1.5) rectangle (-3.5,1.5) node[pos=0.5] {corrupted data domain\\\\\\\\\\} node[pos=0.5] (M) {$X\subtext{corrupted}( \o_j )$\\\footnotesize{(\textbf{input})}};
		\draw [{Circle[length=4pt]}-latex] (M) -- (PL) node [midway, above, sloped] () {$g\subtext{corrupted}\suptext{(CG)MCA}$} node [midway, below, sloped] () {\footnotesize{(\textbf{known})}};
		\draw[draw=black] (+3.5,-2) rectangle (+8,2) node[pos=0.5] {unmodified data domain\\\\\\\\\\\\\\};
		\draw [-latex] (U.south west) -- (PR) node [midway, above, sloped] () {$g\subtext{unmodified}\suptext{(CG)MCA}$} node [midway, below, sloped] () {\footnotesize{(\textbf{known})}};
		\node (ONW) at (O.north west) {};
		\draw[-latex,very thick] (PL.center) to [bend right] node [below] () {least squares$\quad$} (ONW);
%		\fill[fill = white] (U.south west) circle (1pt);
%		\draw[-latex,very thick] (PL.center) to [bend right] node [below] () {least squares$\quad$} (N);
%		\draw[{Circle[length=4pt]}-latex,very thick] (PL) to [bend right] node [below] () {least squares$\quad$} (N);
		\node (USW) at (U.south west) {};
		\fill (USW) circle (2pt);
		\fill[fill = white] (USW) circle (1pt);
	\end{tikzpicture}
\caption{Schematic representation of the least-squares problem \eqref{lsqr_problem} solved for each $j \in \text{inds}\subtext{test}^{ ( d ) }$ for each digit $d$.  Empty circles are unknown points.  The known data point $X\subtext{corrupted}( \o_j )$ and unknown data point $X\subtext{unmodified}( \o_j )$ are mapped closely together in the common domain via the (CG)MCA maps $g\subtext{corrupted}\suptext{(CG)MCA}$ and $g\subtext{unmodified}\suptext{(CG)MCA}$, respectively.  The least-squares procedure (thick arrow) serves as an approximate inversion of $g\subtext{unmodified}\suptext{(CG)MCA}$, recovering $x_{ j , \text{(CG)MCA} }^{ ( d ) }$, our best guess for $X\subtext{unmodified}( \o_j )$.}
\label{fig:lsqr_schematic}
\end{figure}

For each $j \in \text{inds}\subtext{test}^{ ( d ) }$, we find via least squares (using Matlab's \texttt{lsqr})
\begin{subequations}
\label{lsqr_problem}
\bq
x_{ j , \text{MCA} }^{ ( d ) } \in \arg \min_{ x \in \R^{ 784 \times 1 } } \quad \| g\subtext{unmodified}\suptext{MCA}( x ) - g\subtext{corrupted}\suptext{MCA}( X\subtext{corrupted}( \o_j ) ) \|
\eq
and
\bq
x_{ j , \text{CGMCA} }^{ ( d ) } \in \arg \min_{ x \in \R^{ 784 \times 1 } } \quad \| g\subtext{unmodified}\suptext{CGMCA}( x ) - g\subtext{corrupted}\suptext{CGMCA}( X\subtext{corrupted}( \o_j ) ) \|.
\eq
\end{subequations}
Here, $x_{ j , \text{(CG)MCA} }^{ ( d ) }$ is our best guess, in the least-squares sense, at the unmodified image $X\subtext{unmodified}( \o_j )$ that maps as closely as possible under $g\subtext{unmodified}\suptext{(CG)MCA}$ to $g\subtext{corrupted}\suptext{(CG)MCA}( X\subtext{corrupted}( \o_j ) )$, the common-domain representation of $X\subtext{corrupted}( \o_j )$.  The goal of this testing procedure, then, is to take as input a corrupted image $X\subtext{corrupted}( \o_j )$ for some $j$ and find its unmodified counterpart $X\subtext{unmodified}( \o_j )$.  A schematic representation of this least-squares problem can be found in Figure \ref{fig:lsqr_schematic}.

\vspace{0.15in}

%\begin{figure}[h]
%\centering
%	\begin{tikzpicture}[every text node part/.style={align=center}]
%		\node[shape=circle,draw=black] (A) at (0,0) {\\{}common\\{}domain\\{}};
%		\fill (0,-0.75) circle (2pt) node (P) {};
%		\draw[draw=black] (-8,-1.5) rectangle (-3.5,1.5) node[pos=0.5] {corrupted data domain\\\\\\\\\\} node[pos=0.5] (M) {$X\subtext{corrupted}( \o_j )$\\\footnotesize{(\textbf{input})}};
%		\draw [-latex] (M) -- (P) node [midway, above, sloped] () {$g\subtext{corrupted}\suptext{(CG)MCA}$};
%		\draw[draw=black] (+3.5,-1.5) rectangle (+8,1.5) node[pos=0.5] {unmodified data domain\\\\\\\\\\} node[pos=0.5] (N) {$x_{ j , \text{(CG)MCA} }^{ ( d ) }$\\\footnotesize{(\textbf{output})}};
%		\draw [-latex] (N) -- (P) node [midway, above, sloped] () {$g\subtext{unmodified}\suptext{(CG)MCA}$};
%		\draw[-latex,very thick] (P) to [bend right] node [below] () {least squares$\quad$} (N);
%	\end{tikzpicture}
%\caption{Schematic representation of the least-squares problem solved for each $j \in \text{inds}\subtext{test}^{ ( d ) }$ for each digit $d$.  The least-squares procedure serves as an approximate inversion of $g\subtext{unmodified}\suptext{(CG)MCA}$.}
%\label{fig:lsqr_schematic}
%\end{figure}

We note that least squares is a natural choice of inversion process: the fundamental idea of (CG)MCA is to map points closely together in the common domain, so it is natural to select an inversion process based on minimizing distances between common-domain points.

As shown in Figure \ref{fig:mini_montage}, the least-squares problem \eqref{lsqr_problem} produces digits that are discernible, but of low contrast.  In order to create more recognizable visualizations of the least-squares solutions, we form filtered images $\hat{ x }_{ j , \text{(CG)MCA} }^{ ( d ) }$ via Matlab's built-in functions as
\bq
\label{filtered_imgs}
\hat{ x }_{ j , \text{(CG)MCA} }^{ ( d ) } \equiv \texttt{medfilt2}( \texttt{wiener2}( x_{ j , \text{(CG)MCA} }^{ ( d ) } , \texttt{[ 3 3 ]} )
\eq
and evaluate the closeness of each filtered image to the unmodified image $X\subtext{unmodified}( \o_j )$ via Matlab's image similarity metric \texttt{ssim} \cite{ssim}.  Here, $\texttt{wiener2}( \, \cdot \, , \texttt{[ 3 3 ]} )$ is a de-noising filter over $3 \times 3$ neighborhoods \cite{wiener2}, while \texttt{medfilt2} is a contrast-enhancing median filter, also over $3 \times 3$ neighborhoods \cite{medfilt2}, each chosen empirically.

\subsection{Results}\label{sec:results}

\begin{figure}[h!]
\centering
\includegraphics[scale=0.2]{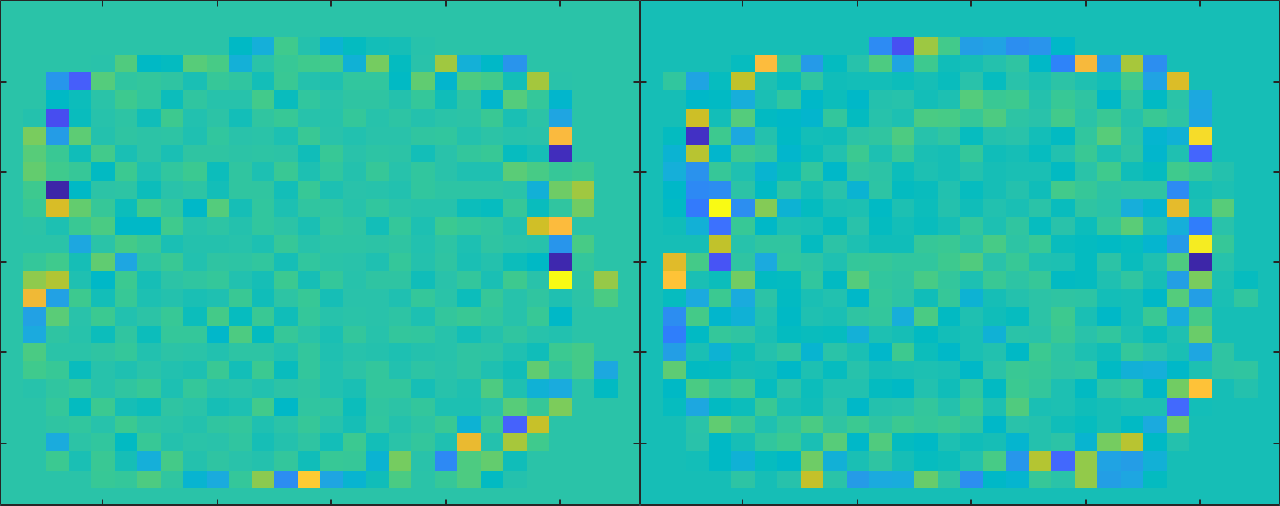}
\caption{The solutions $x_{ j , \text{MCA} }^{ ( 3 ) }$ (left) and $x_{ j , \text{CGMCA} }^{ ( 3 ) }$ (right) of the least-squares problem \eqref{lsqr_problem} for the same ``3'' as in Figure \ref{fig:data_domain_montage}.}
\label{fig:mini_montage}
\end{figure}

Figure \ref{fig:mini_montage} shows images of the solutions $x_{ j , \text{MCA} }^{ ( d ) }$ and $x_{ j , \text{CGMCA} }^{ ( d ) }$ of the least-squares problem \eqref{lsqr_problem} for the same ``3'' shown in Figure \ref{fig:data_domain_montage}.  The ``3'' is visible in the CGMCA solution, but not in the MCA solution.

\begin{figure}[h!]
\centering
\includegraphics[scale=0.2]{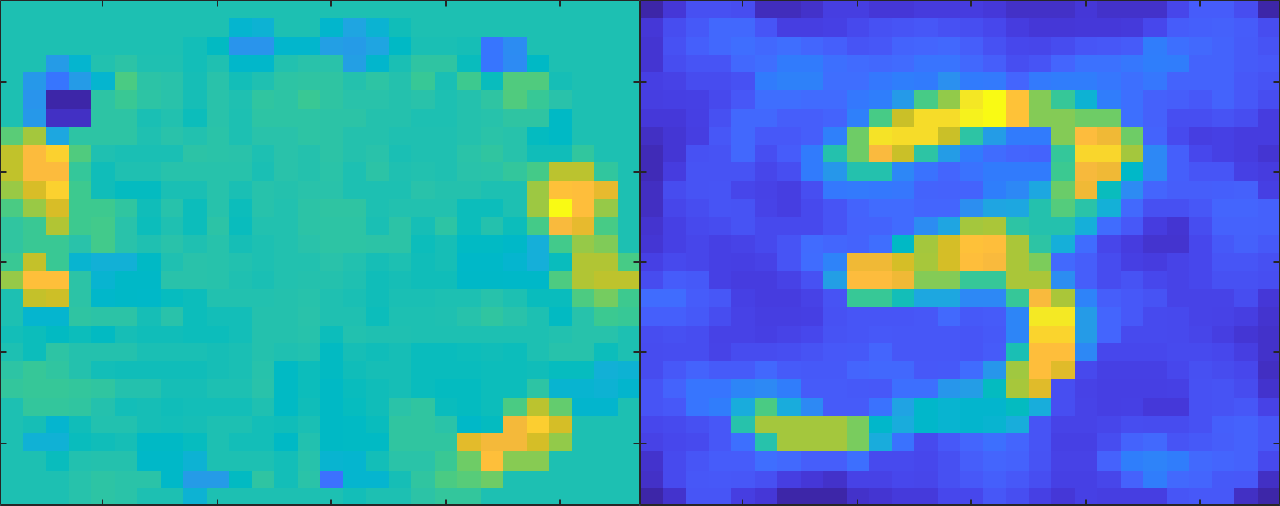}
\caption{The filtered images $\hat{ x }_{ j , \text{MCA} }^{ ( 3 ) }$ (left) and $\hat{ x }_{ j , \text{CGMCA} }^{ ( 3 ) }$ (right) of \eqref{filtered_imgs} for the same ``3'' found in Figures \ref{fig:data_domain_montage} and \ref{fig:mini_montage}.}
\label{fig:enhanced_montage}
\end{figure}

Figure \ref{fig:enhanced_montage} shows the filtered images $\hat{ x }_{ j , \text{MCA} }^{ ( 3 ) }$ and $\hat{ x }_{ j , \text{CGMCA} }^{ ( 3 ) }$ of \eqref{filtered_imgs} for the same ``3'' as in Figures \ref{fig:data_domain_montage} and \ref{fig:mini_montage}.  After identical filtering, the MCA image shows no identifiable ``3'', while the CGMCA image shows a recognizable ``3''.

\begin{figure}[h!]
\centering
\includegraphics[scale=0.35]{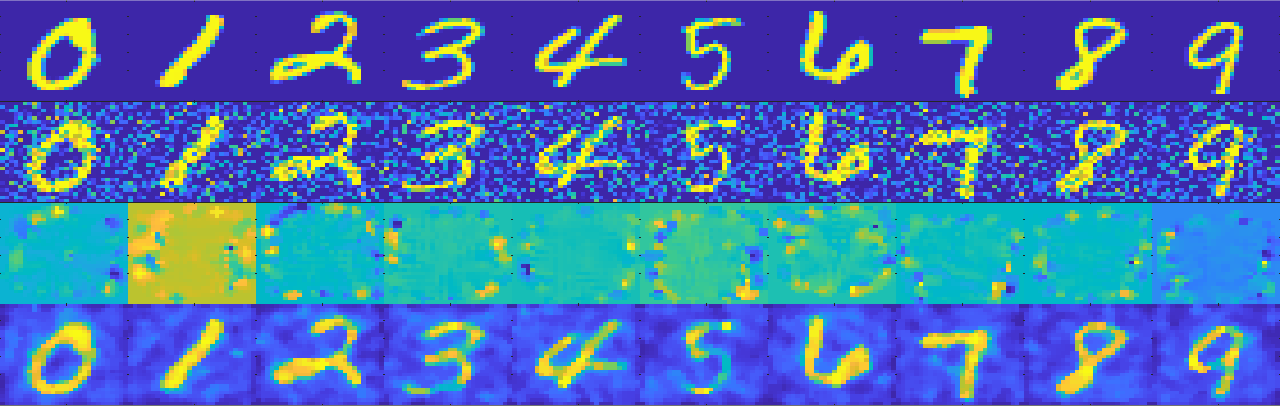}
\caption{Unmodified (first row), corrupted (second row), filtered MCA (third row), and filtered CGMCA (fourth row) example images of each digit; see definitions \eqref{Xc_unmodified_d} and \eqref{filtered_imgs}.}
\label{fig:full_montage}
\end{figure}

For each digit $d$, Figure \ref{fig:full_montage} shows an example element of $\mathcal{ X }\subtext{unmodified}^{ ( d ) }$ (first row) and of $\mathcal{ X }\subtext{corrupted}^{ ( d ) }$ (second row) and an example image of $\hat{ x }_{ j , \text{MCA} }^{ ( d ) }$ (third row) and of $\hat{ x }_{ j , \text{CGMCA} }^{ ( d ) }$ (fourth row); see definitions \eqref{Xc_unmodified_d} and \eqref{filtered_imgs}.

These figures indicate that CGMCA is preserving covariance information that is destroyed by MCA.  In the case of MCA, the inversion process -- here, least squares -- must transform data with an identity covariance (in the common domain) into data with a more appropriate covariance (that of the unmodified data).  In the case of CGMCA, however, the inversion process can simply preserve the common-domain covariance information already enforced by the CGMCA maps.

For each digit $d$, Table \ref{tab:ssim} shows the value of $t$ (chosen empirically) and the average \texttt{ssim} score (higher is better) over each set $\text{inds}\subtext{test}^{ ( d ) }$ after filtering via \eqref{filtered_imgs} and scaling pixel values to be in $[ 0 , 1 ]$.  Here, the \texttt{ssim} values are evaluated by comparing the filtered images \eqref{filtered_imgs} to the true unmodified images $X\subtext{unmodified}( \o_j )$.

\begin{table}[h!]
\begin{center}
\begin{tabular}{ r|cccccccccc } 
 & 0 & 1 & 2 & 3 & 4 & 5 & 6 & 7 & 8 & 9\\
\hline
$t$ & 500 & 500 & 500 & 550 & 550 & 550 & 500 & 500 & 500 & 500 \\
mean \texttt{ssim} (MCA)	& 0.0122 & 0.0037 & 0.0125 & 0.0095 & 0.0122 & 0.0113 & 0.0096 & 0.0088 & 0.0099 & 0.0077 \\
mean \texttt{ssim} (CGMCA)	& 0.1556 & 0.0271 & 0.1293 & 0.1320 & 0.1118 & 0.1413 & 0.1252 & 0.1034 & 0.1055 & 0.0819 \\
\end{tabular}
\caption{\texttt{ssim} values (higher is better) over each digit's entire test set $\text{inds}\subtext{test}^{ ( d ) }$.}
\label{tab:ssim}
\end{center}
\end{table}

From Table \ref{tab:ssim}, we see that the images produced by CGMCA score much higher (usually by a factor of $10$ or more) than those produced by MCA, indicating a higher degree of similarity to the true unmodified images.

\section{Discussion}\label{sec:discussion}

Our key theoretical results are the CGMCA maps \eqref{opt_soln} and their method of computation, as provided in Algorithm \ref{CovGenMCA}.%  These maps are capable of encoding meaningful variational information not able to be captured by the identity covariance constraint of CGMCA's predecessor, MCA.

Our key numerical result is summarized in Figure \ref{fig:full_montage}, which demonstrates our claim that CGMCA's maps are capable of meaningfully encoding variational information not able to be captured by the identity covariance constraint of CGMCA's predecessor, MCA.  Further evidence of this claim is provided in Table \ref{tab:ssim}, which quantitatively compares entire test sets for each digit $d$ under MCA and CGMCA and shows that CGMCA produces images more similar to the true images than those produced by MCA.

In contrast to the identity matrix covariance constraint $C_i C_i\trans \equiv I_k$ of \cite{MCA}, the generalized covariance constraint \eqref{original_constraint_cov} may be used to encode both
\begin{enumerate}[labelindent=0.25in, style=multiline, leftmargin=0.75in]
	\item[\textbf{(E1)}] uncertainty information about the ability of the measurement process underlying some data domain to provide information about elements of that data domain, and
	\item[\textbf{(E2)}] variational information about the elements of a data domain, especially when those elements are defined probabilistically.
\end{enumerate} 
While other encodings are possible, we expect the encodings \textbf{(E1)} and \textbf{(E2)} to have potential utility in any data fusion or transfer learning applications whose objects of study are statistical or probabilistic.

\section{Conclusion}

In order to encode additional statistical information in data fusion and transfer learning contexts, this paper introduced a generalized covariance constraint for MCA.  The resulting covariance-generalized optimization problem was solved in closed form, and an algorithm for the computation of this solution was provided.  This generalized technique is called CGMCA, and was demonstrated via numerical experiments to be capable of encoding meaningful variational information not able to be captured by the identity covariance constraint of CGMCA's predecessor, MCA.

We note that the CGMCA technique has the potential to be useful in any data fusion or transfer learning application in which encodings afforded by the generalized covariance constraint \eqref{original_constraint_cov} -- such as \textbf{(E1)} and \textbf{(E2)} of Section \ref{sec:discussion} -- are desired.

For an example of the application of this paper's theory to the challenge of characterizing microtexture regions in titanium, see \cite{Homa_CGMCA}.

\section*{Acknowledgments}

The authors would like to acknowledge support from the Air Force Research Laboratory (AFRL) through contract FA8650-19-F-5230 and from the Air Force Office of Scientific Research (AFOSR) through grant 21RXCOR037 under the Dynamic Data and Information Processing (DDIP) program.

The authors would also like to acknowledge Laura Homa for helpful discussions involving the numerical experiments.

\end{document}